\documentclass[10pt,twocolumn,letterpaper]{article}
\usepackage{cvpr}
\usepackage{times}
\usepackage{epsfig}
\usepackage{graphicx}
\usepackage{amsmath}
\usepackage{amssymb}
\usepackage{subcaption}

\usepackage{algorithm2e}
\usepackage{tabularx}

\usepackage[pagebackref=true,breaklinks=true,letterpaper=true,colorlinks,bookmarks=false]{hyperref}

\cvprfinalcopy % *** Uncomment this line for the final submission

\def\cvprPaperID{6942} % *** Enter the CVPR Paper ID here

\ifcvprfinal\fi
\begin{document}

%%%%%%%%% TITLE
\title{Enhancing Generic Segmentation with Learned Region Representations}
% \author{Or Isaacs\thanks{Both authors contributed equally}\\
% Technion\\
% Haifa, Israel\\
% {\tt\small sisaacs@cs.technion.ac.il}
% \and
% Oran Shayer\footnotemark[1]\\
% Technion\\
% Haifa, Israel\\
% {\tt\small oran.sh@gmail.com}
% \and
% Michael Lindenbaum\\
% Technion\\
% Haifa, Israel\\
% {\tt\small mic@cs.technion.ac.il}
% }
\author{Or Isaacs\thanks{Both authors contributed equally}\hspace{1cm} Oran Shayer\footnotemark[1]\hspace{1cm} Michael Lindenbaum\\
Computer Science, Technion - Israel Institute of Technology\\
{
\tt\small orisaacs@gmail.com} \hspace{0.8cm} 
{\tt\small oran.sh@gmail.com}
\hspace{0.8cm} 
{\tt\small mic@cs.technion.ac.il}
}

\maketitle
%\thispagestyle{empty}

%%%%%%%%% ABSTRACT
\begin{abstract}
Current successful approaches for generic (non-semantic) segmentation rely mostly on edge detection and have leveraged the strengths of deep learning mainly by improving the edge detection stage in the algorithmic pipeline. This is in contrast to semantic and instance segmentation, where DNNs are applied directly to generate pixel-wise segment representations.
We propose a new method for learning a pixel-wise representation that reflects segment relatedness. This representation is combined with an edge map to yield a new segmentation algorithm.
We show that the representations themselves achieve state-of-the-art segment similarity scores. Moreover, the proposed combined segmentation algorithm provides results that are either state of the art or improve upon it, for most quality measures. 

\end{abstract}

\section{Introduction}

Generic segmentation is the well-studied task of partitioning an image into parts that correspond to objects for which no prior information is available. Deep learning approaches to this task thus far have been indirect and relied on 
% a high quality 
edge detection. The COB algorithm \cite{cob}, for example, uses a learned edge detector to suggest a high-quality contour map, and then creates a segmentation hierarchy from it using the oriented watershed transform \cite{gPbucm}.

In this work, we follow the approach used for semantic segmentation: learning pixel-wise representations that capture region and segment characteristics. However, we apply this approach for the first time to the generic (non-semantic) segmentation task. This paper focuses on creating such representations, along with combining them with edge-based information to a generic segmentation algorithm improving the current state of the art.

Deep learning has been successfully used in a supervised regime, where the network is learned end-to-end on supervised tasks (e.g., classification \cite{resnet}, object detection \cite{yolo}, semantic segmentation \cite{deeplab}, or edge detection \cite{hed}). Generic segmentation, however, cannot be formulated as such. Unlike semantic segmentation, the properties of regions inside segments of new ("test") images are not well specified with respect to the regions or the objects in a training set, and therefore, the direct classification approach does not apply.

\begin{figure}[t]
\centering
 \includegraphics[width=0.4\textwidth]{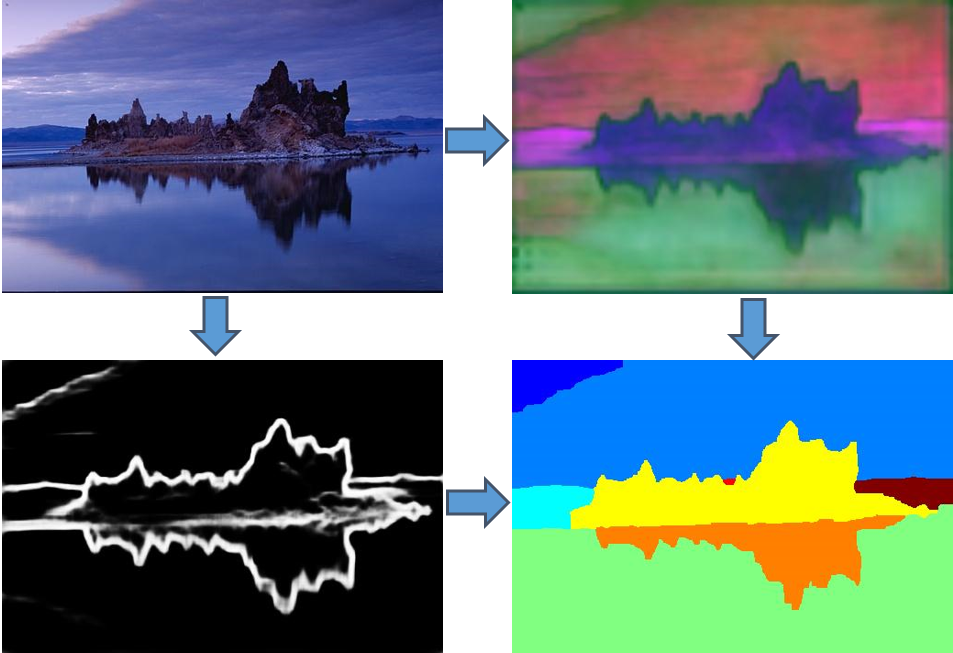}
\caption{The algorithm flow: the proposed region representation (top right) and the edges (bottom left) of the input image (top left) are extracted and then combined to the resulting segmentation (bottom right). }
\label{fig:pipeline}
\end{figure}

The common task of face verification (\cite{deepface}, \cite{facenet}, \cite{face_veldadi}) shares this difficulty. Even if we learn on thousands of labeled faces, there may be  millions of unseen faces that must be handled. To succeed, we may 
% must be able to 
learn a model, or representation, that captures properties capable of distinguishing between different faces, even those not encountered in training.
Similarly, in generic segmentation, the images to be partitioned might contain objects not seen in the training set.
The problem is further complicated by the fact that the annotated segments have unknown semantic meaning (i.e., we do not know what objects or parts are marked).
%The problem is further complicated by two factors: the annotated segments have unknown semantic meaning (i.e., we do not know what objects or parts are marked), and the number of segments in each image is also unknown.

Thus,  we learn pixel-wise representations that express segment relatedness. That is, the representations are grouped together in representation space for pixels of the same segment, and kept further apart for pixels from different segments.
%
% In this paper, we propose to learn such representations by 
We use a supervised algorithm which follows the principles of the DeepFace algorithm \cite{deepface} but addresses the differences between the tasks. % segmentation and face classification. 
We compare it to the more common approach for deep representation learning (triplet loss \cite{hoffer2015triplet}), and test it quantitatively and visually. 

Our Boundaries and Region Representation Fusion (BRRF) algorithm combines the learned representation with edges (from an off-the-shelf edge detector), as illustrated in fig. \ref{fig:pipeline}.

Our contributions in this paper are as follows:
\begin{enumerate}
\item We present (the first) pixel-wise representation for generic segmentation. This representation captures segmentation properties and performs better than previous methods on a pixel pair classification task.
% \item We implement a new deep learning approach for learning such pixel-wise segment representations.
\item We present a new segmentation algorithm that  uses both the proposed pixel-wise representations and the traditionally used edge detection. This % combined segmentation 
algorithm achieves excellent results, and for some quality measures, significantly improves the state of the art. 
% either achieves the state of the art or improves it (depending on the quality measure). 
\end{enumerate}

\section{Related Work}

\subsection{Generic segmentation}\label{intro:GenSeg}
Generic segmentation has seen a broad range of approaches and methods. 
% Here we mention only a few examples. 
Earlier methods (e.g., \cite{meanshift}) 
rely on clustering of local features.
% The features are considered as random instances drawn from a distribution, and the goal is to find the modes of this distribution by iteratively applying mean shift filtering. Pixels that converge to the same mode are grouped together. 
Modern methods often rely on graph representations, in which pixels or other image elements are represented by graph nodes, and weights on the edges may represent the similarity or dissimilarity between them. Then, segmentation is carried out by cutting the graph into dissimilar parts % The intuitive idea of dividing the image into two parts that are most dissimilar translates into finding the minimal cut in this similarity graph.
\cite{minimal_cut-boykov, ncuts}.  % criterion  guarantees that no group is too small, and the problem of finding the minimal normalized cut is elegantly solved using generalized eigenvectors. 
% The weights on the edges can also represent the dissimilarity between the nodes. 
The bottom-up watershed algorithm \cite{watershed}, for example, merges the two elements with the least dissimilar nodes at each iteration.
% smallest dissimilarity values between them. 
% A merging criterion that is sensitive to the typical dissimilarity within the merged segments and to their sizes significantly improves the results  
See also \cite{felzhutt}. 
% for an improved merging criterion.

The OWT-UCM algorithm \cite{gPbucm} uses edge detection to get reliable dissimilarities, and an oriented watershed transform to transform the graph into a hierarchical region tree. % It then modifies the weights so that thresholding them yields a set of closed curves and well-specified segmentation. Using different thresholds gives the hierarchy. 
Combining multiple scales further improves performance \cite{mcg}. This approach, coupled with a CNN edge detector, achieves the current state-of-the-art \cite{rcf-19}.

An efficient and elegant way to represent the hierarchical region tree is through an Ultrametric Contour Map (UCM) \cite{ucmMap}, also known as a Saliency Map \cite{watershed}. The UCM assigns a value to each contour in the image so that a contour that persists longer in the hierarchy gets a higher value.
% than a contour that persists less. 
Thresholding the map provides a specific segmentation.
%
% \subsection{Semantic segmentation}

% Deep semantic segmentation, on the other hand,  builds on the ability of fully convolutional NNs to identify the pixels associated with particular categories \cite{fcn}. 
% Skip connections, deconvolution, and dilated (atrous) convolutions were used to maintain and improve output resolution \cite{fcn,deconvnet,deeplab}. Pyramid pooling \cite{psp} and an encoder-decoder architecture \cite{large_kernel_matters} were used to capture larger context.

%A combination with CRF was used to get better localized boundaries\cite{fc-crf}. 

\subsection{Representation learning}

%siamese
Representations can be learned explicitly using a Siamese network \cite{siamese2005, koch15, siamese2006}. An example is a pair of inputs, either tagged as same (positive example) or not same (negative example). Both inputs are mapped to a representation through neural networks whose weights are tied. The networks are trained to minimize the distance in representation space between positive examples and increase the distance between negative examples.
The learning can also be made through a triplet setting \cite{hoffer2015triplet,facenet}, where an example is a triplet of inputs. The first two inputs are positive and negative, respectively, relative to the third one.
%An example may also be a triplet of inputs \cite{hoffer2015triplet,facenet}, where the first two inputs are positive and negative, respectively, relative to the third one.

% supervised task
The representation can also be learned implicitly by learning a supervised task. 
% In a supervised setting, 
The last layer of a network can be regarded as a classifier, while the rest of the network generates a representation that is fed to this classifier \cite{decaf}. These representations can be used later on to distinguish between unseen classes \cite{deepface}, or for transfer learning \cite{decaf}.

The closest work to the representation part of this paper is Patch2Vec \cite{p2v}, which %(explicitly) 
learns an embedding for image patches by training on triplets tagged according to the segmentation. Our approach differs by using implicit learning and allowing a larger context from the full image.
%DEL \cite{del} present representation learning for generic segmentation, but for super-pixels. They then use representation distance as the dissimilarity measure according to which they merge. Constraining the representation to super-pixels injects inherent constraints, and their agglomerative process is quite simple, relaying solely on super-pixel representation distance.
Superpixel representations for generic segmentation are learned in \cite{del}. This work differs from our per-pixel representation. As we will show later on, being pixel-based is important because it allows us to give consideration to edges, and construct the effective similarity that is later used. Besides, it is learned differently, using contrastive loss in a Siamese setting, and leads to segmentation results that are not as good as those achieved in recent works (including ours).

\section{Representing Non-Semantic Segmentation}
\label{representation}

One well-known strength of neural networks is their ability to capture both low-level and high-level features of images, creating powerful 
% and useful 
representations \cite{zeiler2014visualizing, bengio2013representation}. 
In this work, we focus on segmentation-related representations and harness this strength to provide a new pixel-wise representation. This learned representation should capture the segment properties of each pixel, so that representations associated with pixels that belong to the same segment are close in representation space, and their cluster is farther away from clusters representing different segments.
This representation is a pixel-wise $N$-dimensional vector (thus, the full image representation, denoted $R$, is an $H\times W\times N$ tensor). 

Learned classifiers have been used in the context of semantic (model-based) segmentation. Learning a classifier directly is possible for the semantic segmentation task because every pixel is associated with a clear label: either a specific category or background.  This is not the case with generic segmentation, where object category labels are not available during learning and are not important at inference. Moreover, at inference, the categories associated with the segments are not necessarily those used in training.

\subsection{Learning the representation}\label{Learning-rep}

%\subsubsection{Explicit learning -- Siamese or triplet loss}\label{explicit}
A pixel-wise representation for generic segmentation can be learned explicitly by minimizing a triplet loss over triplets \cite{hoffer2015triplet,facenet}.
%A triplet input consists of an anchor, a positive example (same segment) with respect to the anchor, and a negative example (not same) relative to the anchor. We then minimize the following triplet loss:
%\begin{multline}
%  \mathcal{L}_{triplet} = \sum_{i} L(R_i, R_j^+, R_k^-), \ \ \ where \\
%\end{align}
% where
%\begin{align}
%L(R_i, R_j^+, R_k^-) = \left[d(R_i, R_j^+)\right]^2 + max\left[0, m-d(R_i, R_k^-)\right]
%\end{multline}
While this approach has proved beneficial for high-level image representations or patches, it has not been explored on tasks which provide structured outputs such as segmentation. We shall compare to this approach later in this paper.

\begin{figure}[t]
\centering

\begin{subfigure}[b]{0.15\textwidth}
  \includegraphics[width=\textwidth]{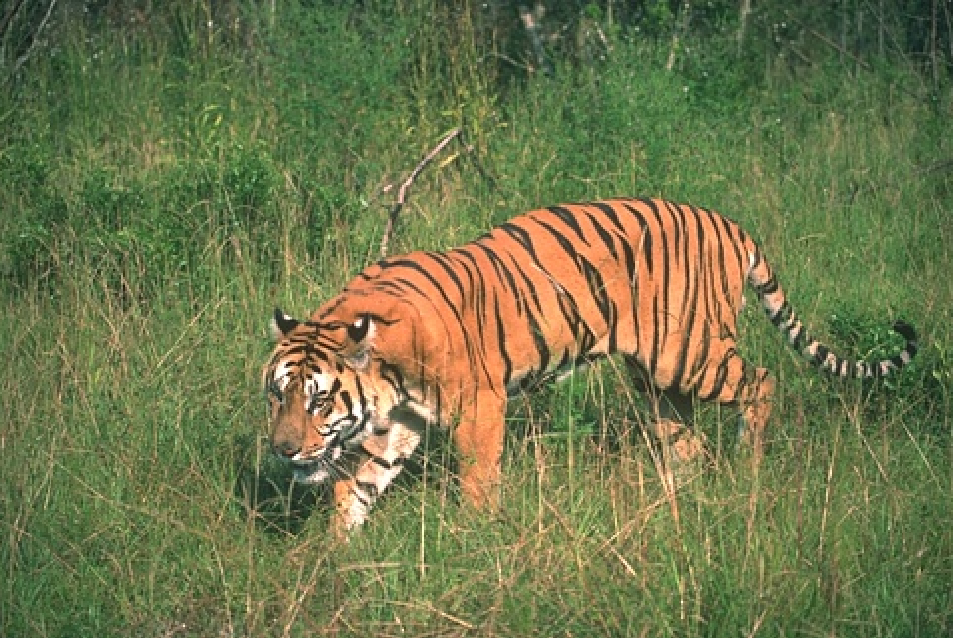}
\end{subfigure}
\begin{subfigure}[b]{0.15\textwidth}
  \includegraphics[width=\textwidth]{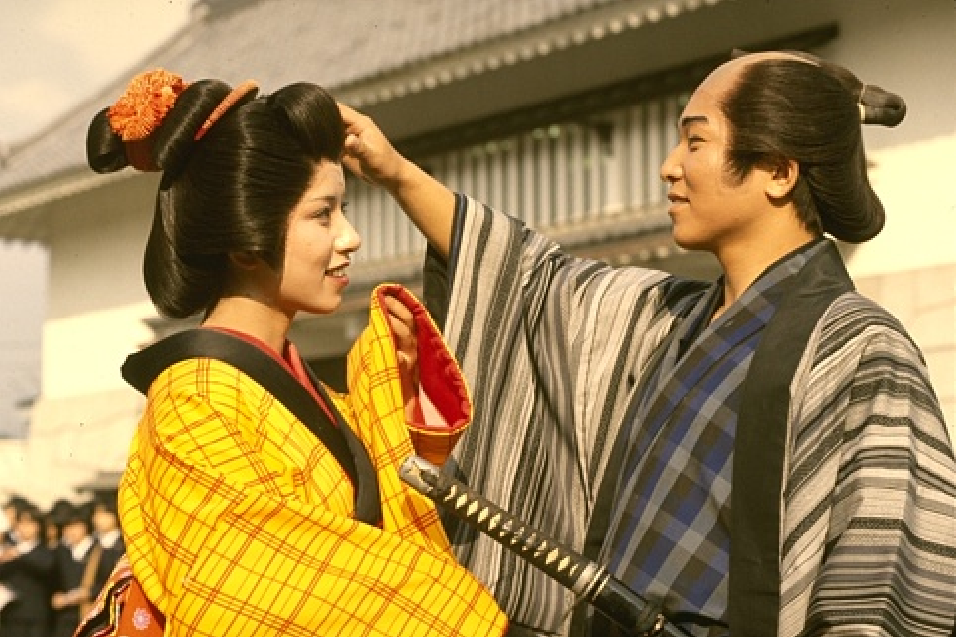}
\end{subfigure}
\begin{subfigure}[b]{0.15\textwidth}
  \includegraphics[width=\textwidth]{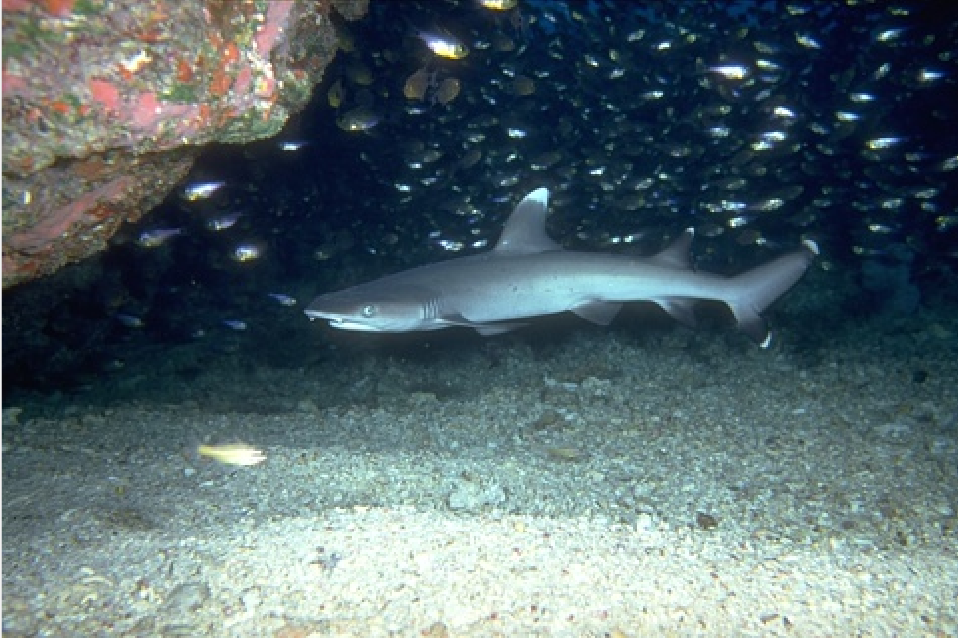}
\end{subfigure}
\vspace{1 mm}

\begin{subfigure}[b]{0.15\textwidth}
  \includegraphics[width=\textwidth]{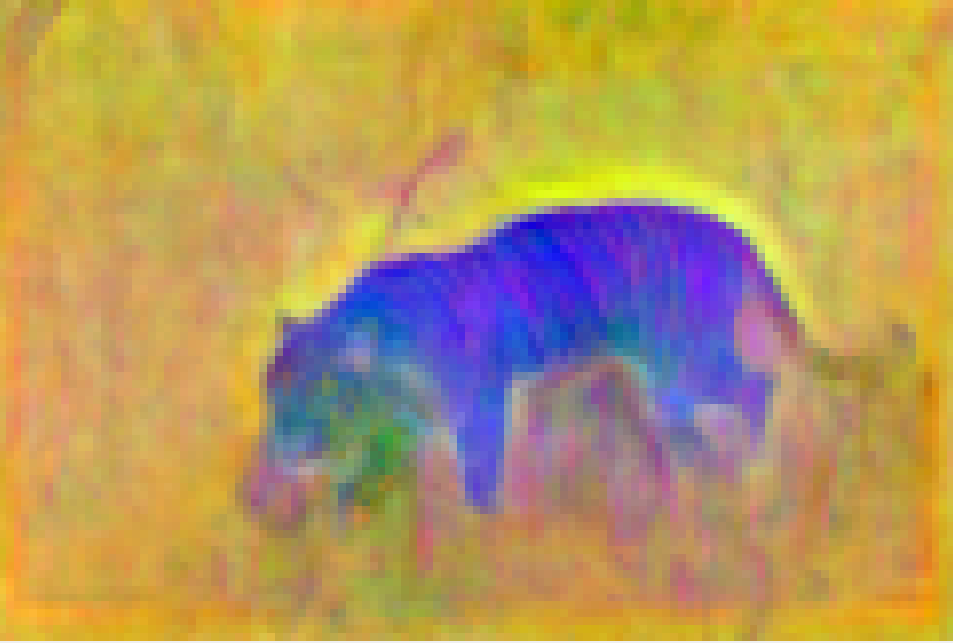}
\end{subfigure}
\begin{subfigure}[b]{0.15\textwidth}
  \includegraphics[width=\textwidth]{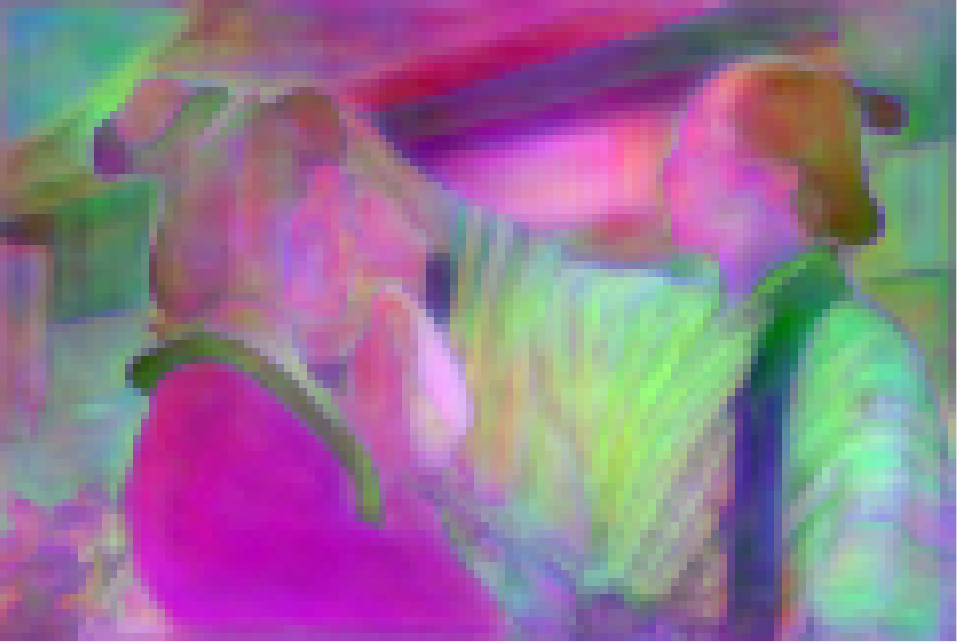}
\end{subfigure}
\begin{subfigure}[b]{0.15\textwidth}
  \includegraphics[width=\textwidth]{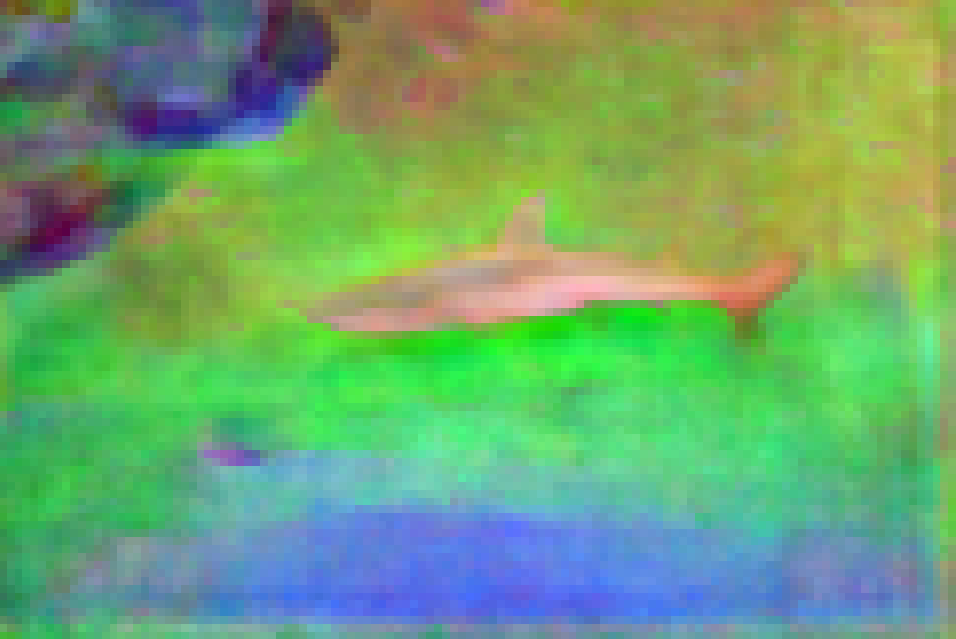}
\end{subfigure}
\vspace{1 mm}

\begin{subfigure}[b]{0.15\textwidth}
  \includegraphics[width=\textwidth]{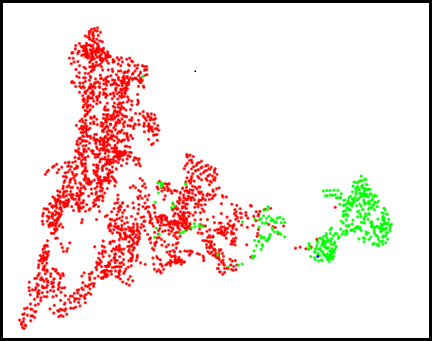}
\end{subfigure}
\begin{subfigure}[b]{0.15\textwidth}
  \includegraphics[width=\textwidth]{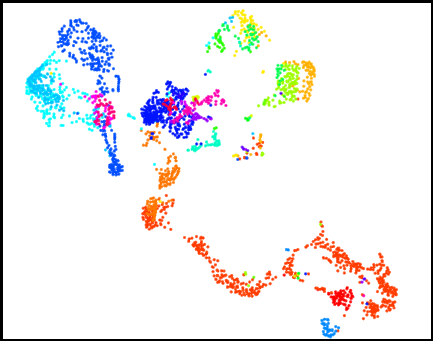}
\end{subfigure}
\begin{subfigure}[b]{0.15\textwidth}
  \includegraphics[width=\textwidth]{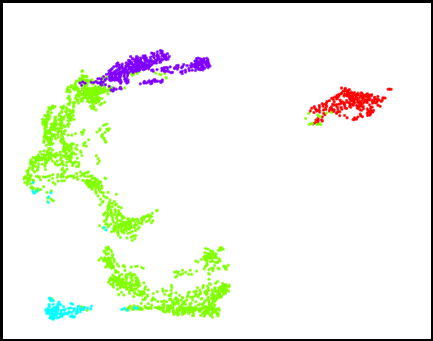}
\end{subfigure}
\vspace{1 mm}

\begin{subfigure}[b]{0.15\textwidth}
  \includegraphics[width=\textwidth]{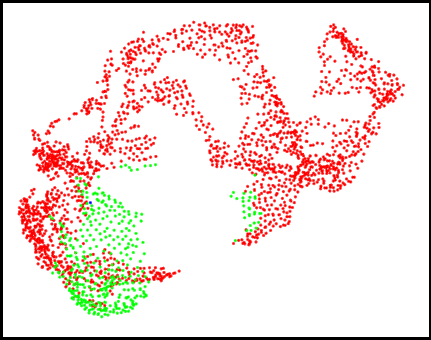}
\end{subfigure}
\begin{subfigure}[b]{0.15\textwidth}
  \includegraphics[width=\textwidth]{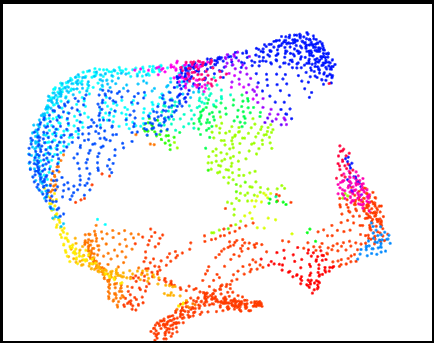}
\end{subfigure}
\begin{subfigure}[b]{0.15\textwidth}
  \includegraphics[width=\textwidth]{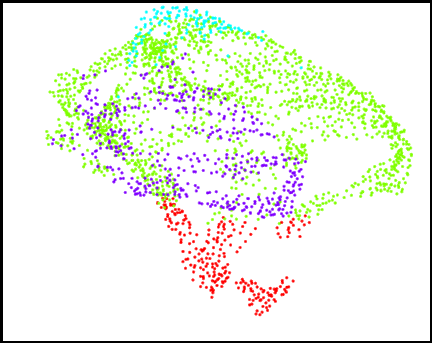}
\end{subfigure}
\vspace{1 mm}

\caption{In each column, top to bottom: original image, representation space virtual colors, t-SNE of our representation, and t-SNE of the representation from a network trained for semantic segmentation. In t-SNE plots, points of the same color belong to the same segment. Notice how areas such as the tiger or the woman's kimono are nearly uniform in color, indicating that those pixels are close in representation space. 
% In addition, notice the sharp boundaries in the virtual colors, evidence that the representation is boundary aware and represents the pixel accordingly. 
Note that the t-SNE plots formed from our representation are better separated than the alternative.}\label{pca_tsne}
\end{figure}

%\subsubsection{Implicit learning}
Representations can also be learned implicitly, by training a network on a related high-level task, and afterwards using the representation from the last hidden layer for the original task. However, using such a learning approach for generic segmentation related representations is not straightforward. To formalize such a high-level task, like classification, the labels must have some semantic meaning. % in order for them to be used for classification.
% old - changes 210220
% For example, in face recognition in the wild (face verification), the categories (face identities) to be classified at test time are usually different and more numerous than those available for training. In the DeepFace approach \cite{deepface}, 
% new
For example, in the DeepFace approach \cite{deepface} for face verification in the wild,  
% the training algorithm learns a face classification task using examples of $K$ (over 4000)  face identity classes.
an $L$-layer classification network is trained with $K$ ($>4000$) face identity classes, % where the $L$th layer is the final linear classification layer.
and the $N$-dimensional response  from layer $L-1$
% , denoted as $f$ in the original work, 
is used as the representation of the input image. 

Training with cross-entropy inherently forms clusters of face images belonging to the same identity \cite{decaf}.
With sufficiently large number of face images, the network generalizes well and generates well-clustered representations even for new images of unseen categories. % These representations were used successfully to separate and classify new unseen faces. 
This usage is reminiscent of our generic segmentation task, where the aim is to separate pixels that belong to different segments not seen in training. We 
% propose to 
adopt this approach for learning a representation for generic segmentation. However, some differences need to be addressed. First, we need a representation for each image pixel rather than a single representation for a full image. We therefore use a fully convolutional network that  outputs an $N$-dimensional vector for every output pixel.

\begin{figure}[b]
\centering
\includegraphics[width=0.48\textwidth]{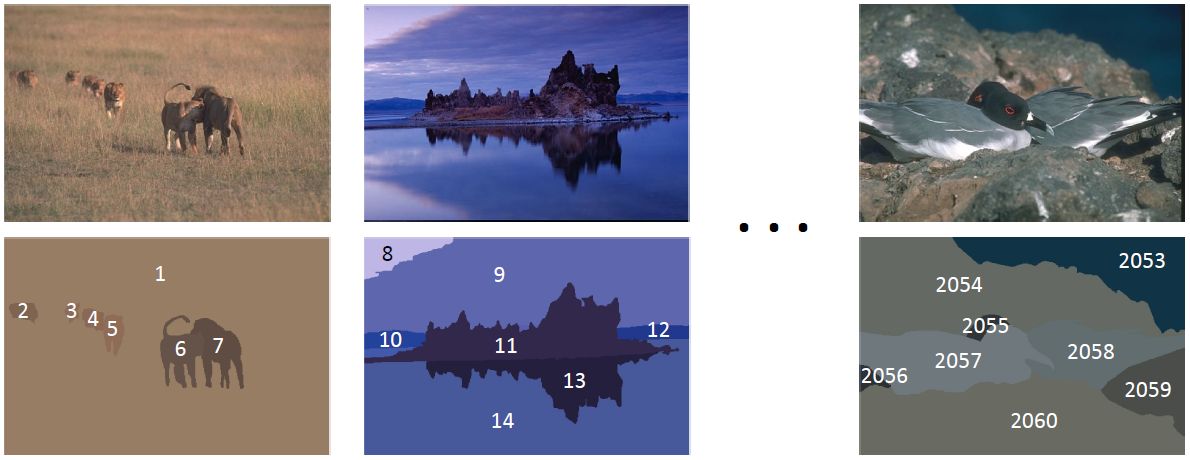}
\caption{Our labeling process. We assign a unique label $l_k$ to each segment in every training image.}\label{fig:labeling}
\end{figure}

A more fundamental difference is that choosing the training labels is not straightforward.  Pixels in segmentation examples are assigned labels depending on the segment they belong to, but unlike face identities, the labels associated with different segments are not meaningful in the sense that they are not associated with object categories or even with appearance types. 
%Segments in different images, for example, may correspond to the same object category (e.g. a horse), but this information is not available for training. 
This creates an ambiguity in the given labeling. For example, in a certain image, label \#$1$ refers to a horse, and in a different image, label \#$1$ refers to the sky.
To address this problem, we consider the set of segments from all images in the training set as different categories. That is, we assign a unique label $l_{k}$ to all pixels in the  i-th segment of the $j$th image ($s_{ij}$), a label that no other pixel in another segment or image is assigned. A visualization of the labeling process can be seen in Fig. \ref{fig:labeling}.

% This paragraph will describe the limitation of predicted classes in each image during training
However, the use of arbitrary categories leads to several difficulties. Two segments of different images may correspond to the same object category and may be very similar (e.g., two segments containing blue sky) but they are considered to be of different classes. Because the two segments have essentially the same characteristics, training a network to discriminate between them would lead to representations that rely on small differences in their properties or on arbitrary properties (e.g., location in the image), both leading to poor generalization. To overcome this difficulty, we modify the training process: when training on a particular image, we  limit the possible predicted classes only to those that  correspond to segments in this image, and not to the segments in the entire training set. 

The fully convolutional NN is trained as a standard pixel-wise classification task which, when successful, will classify each pixel to the label of its segment. In that case, the network will have learned representations which are well-clustered for pixels in the same segment, and different for pixels in different segments. We denote the representation of the $i$-th pixel by  $R_i$. The distance $||R_i - R_j||$ between two pixels should reflect the segment relatedness.

\subsection{The representation learning network}\label{repnet}

\begin{figure}[!h]
\centering
\includegraphics[width=0.45\textwidth]{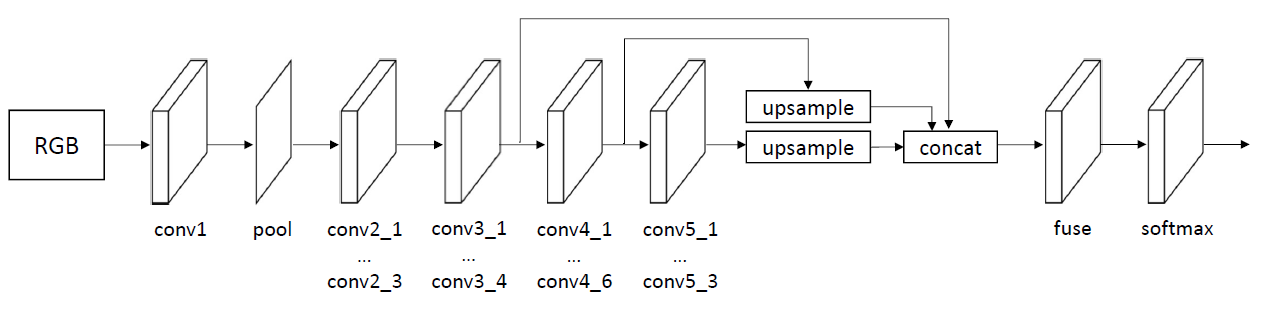}
\caption{Our suggested RepNet architecture.}
\label{repnet-arch}
\end{figure}

The network architecture we use for the representation learning is a modified version of ResNet-50 \cite{resnet}. We make use of layers $conv1$ through $conv5\_3$. To increase the spatial output resolution, we adopt two common approaches: first, atrous (or dilated) convolutions \cite{deeplab} are used throughout layers $conv5\_x$. Second, we use skip connections of layers $conv3\_4, conv4\_6$ and concatenate them with layer $conv5\_3$, upsampling all to the resolution of $conv3\_4$, which is 4 times smaller than the original input resolution.
The concatenated layers pass through a final $fuse$ residual layer, to get a final feature depth of $512$ per pixel. Fig. \ref{repnet-arch} illustrates the network architecture (referred to as RepNet).

In general, the huge pixel-wise classification task (thousands of classes for each pixel) is a major hardware bottleneck that limits our resolution upsampling capabilities. While semantic segmentation task architectures are able to upsample to the original input resolution (the number of classes  are in the range of tens), we were limited to a resolution of $\frac{1}{4}\times$ of the original input image. The upsampling is done with bilinear interpolation. We found that here, deconvolution-based upsampling did not improve performance.

The network was trained using a weighted cross-entropy loss. To improve segment separation, we increased the weight of the loss associated with pixels close to the boundary (closer than $d$) by a factor of $w_b$.
The proposed pixel-wise representation is taken from the final layer before softmax, which we refer to as $fuse$.

\subsection{Visualizing the representations}\label{visualization}

We suggest two visualization options:
\begin{enumerate}
\item \textbf{Representation space virtual colors --} We project the $N$-dimensional representations on their three principal components (calculated using the PCA of all representations). The three-dimensional vector of projections is visualized as an RGB image. We expect pixels in the same segment to have similar projections and similar color. Fig. \ref{pca_tsne} shows this is indeed the case.
\item 
\textbf{t-SNE scatter diagram --} Intuition about the representations can also be gained by using t-Distributed Stochastic Neighbor Embedding (t-SNE, \cite{t-SNE}) to embed them in a 2D space. We expect a separation between points belonging to different segments in the embedded space, and compare the separation to that obtained by a network with the same architecture but trained for semantic segmentation. See  Fig. \ref{pca_tsne}.
\end{enumerate}
Note that the representation changes sharply along boundaries (Fig. \ref{pca_tsne}). This behavior is typical to nonlinear filters (e.g., bilateral filter) and therefore indicates that the representation at a pixel depends mostly on image values associated with the segment containing the pixel and not on image values at nearby pixels outside this segment. That is, the representation describes the segment and is not a simple 
texture description associated with a uniform neighborhood.

\section{A Hierarchical Segmentation Algorithm}
\label{algorithm}

Many modern and successful hierarchical segmentation algorithms work by agglomerating image elements, based on high-quality edge detection results \cite{rcf-19,cob,gPbucm}. 
% In particular, current state of the art results are obtained using deep edge detectors that combine side outputs in multiple stages of a CNN, which was pre-trained on object detection.

To demonstrate the utility of the proposed representation we incorporate it in the agglomerative approach. We use the representation twice. First, for recalculating the dissimilarity value for a pair of segments, and then for re-ranking these values using region context. 
%
%Indeed, adding the representation improves the segmentation over the current state of the art with respect to some of the quality measures. 
%
% To that end we shall construct a scale dependent classifier.   
%
% We propose utilizing our representation along an edge detection algorithm in the process of hierarchical image segmentation. This is achieved by learning a classifier for segment pairs that will output a likelihood for them to belong to the same semantic object and using it to iteratively combine image segments.
Our suggested Boundaries and Region Representation Fusion (BRRF) algorithm for hierarchical image segmentation consists of three stages: 
\begin{enumerate}
\item Initialization: Creating an initial oversegmentation and calculating the proposed region representations. 
\item Using a classifier to calculate an augmented pair dissimilarity for every two neighboring segments. 
\item Iteratively, merging pairs of segments and recalculating dissimilarities using region context. 
\end{enumerate}

\subsection{Algorithm initialization}\label{alg-init}
We obtain an initial oversegmentation using the oriented edge detection results and the watershed algorithm \cite{gPbucm, cob}. We remove small superpixels (SPs) by iteratively combining every SP smaller than 32 pixels with its most similar neighbor - the one with the lowest average edge strength in their shared boundary. This is a common practice in agglomerative segmentation and is necessary here due to the representations being inaccurate for very small SPs.

Independently, we calculate the representation as described in Section \ref{repnet}. To improve performance on small segments, we upsample the representation, through bilinear interpolation, to half of the original image resolution. 
For computational 
% and practical 
reasons, we reduce the representation's dimension to 9 using PCA (calculated separately for this
% over each 
image).
This transformation is effective
% plausible 
because representations in a specific image lie in a low dimension subspace.
% , compared to the full dataset.
% lie in is a subspace of the more general representation space found during the representation network training over the whole dataset.

% \subsection{Pair score calculation}
% To incorporate the representation into the agglomerative process, we use a dissimilarity score using a learned function which is based on the learned representation, edge strength, and some additional properties, describing the segments' geometry and raw color. This dissimilarity score, denoted $pair\_score$, is calculated for every pair of neighbouring segments. A higher score indicates that the two neighboring segments are more likely to belong to different (semantic) object. We  elaborate on the classifier in sec. \ref{classifier}.
\subsection{Pair dissimilarity calculation}

To incorporate the representation into the agglomerative process,
% and to add additional helpful information 
we use a learned dissimilarity function which depends on the representation, edge strength, and some additional properties that describe the segments' geometry and raw color. This dissimilarity, denoted $Pair\_Dissim$, is calculated for every pair of neighbouring segments. A higher dissimilarity indicates that the two segments are more likely to belong to different (semantic) objects. We  elaborate on learning the dissimilarity  in Section \ref{classifier}.

\subsection{The iterative merging stage}

This stage is iterative. At every iteration, we merge two segments into one and then recalculate the dissimilarity between this merged segment and its neighbors. 
%This process creates a hierarchy.

The merged segments may be chosen as the pair associated with the lowest $Pair\_Dissim$ in the current hierarchy. This works well (see Section \ref{exp:features}), but an improved decision is achieved by relying also on a context-based clustering test, which uses the representation as well.

The Silhouette score \cite{silhouette} is an unsupervised measure of cluster separation. It calculates the dissimilarities between elements contained in one cluster and other elements in other clusters, and compares these dissimilarities to the dissimilarity within the cluster. It achieves a high score when the former is high and the latter is low, indicating that the clusters are well separated.
Inspired by the Silhouette score, we consider a segment pair and test whether the union of the two segments is a well separated cluster. 

\begin{figure}[t]
\centering
 \includegraphics[width=0.33\textwidth]{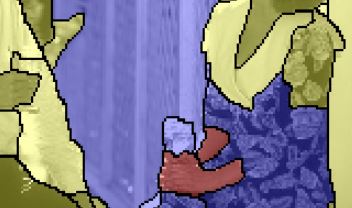}
\caption{The Silhouette score calculation. The  merged segment candidate $S$ (red), the union of its neighboring segments $S_{n}$ (blue), and process irrelevant segments (yellow).}
\label{rerankExample}
\end{figure}

Denote the candidate merged segment by $S$,  and the union of all its neighboring segments (in the current segmentation), by $S_n$; see an example in Fig. \ref{rerankExample}. Let $d(R_i, R_j)$ be the Euclidean distance between the representations $R_i, R_j$, corresponding to pixels $i,j$ respectively. Then, the dissimilarities $a(i), b(i)$ associated with pixel  $i \in S$, and the corresponding Silhouette score are:
\begin{align}
\begin{split}
  &a(i) = \dfrac{1}{|S|-1} \sum_{j \in S, j \neq i} d(R_i,R_j)\\
  &b(i) = \dfrac{1}{|S_{n}|} \sum_{j \in S_{n}} d(R_i,R_j)\\
%  &s(i) = \dfrac{b(i)-a(i)}{max\{a(i),b(i)\}} \\
  &Sil\_Score = \dfrac{1}{|S|} \sum_{i \in S} \dfrac{b(i)-a(i)}{max\{a(i),b(i)\}} 
\end{split}
\end{align}
For large segments, only a subset of the segment pixels are used for both $S$ and $S_n$; see Section \ref{alg-filt}.

This criterion is not accurate enough to be used by itself, as an alternative to the {\em Pair\_Dissim}. This holds especially in the beginning of the merging process, when the segments are small, and are typically not very dissimilar to their neighbors. 
%In addition, applying it to all candidate pairs is computationally expensive.
Therefore, we do not calculate the silhouette score until we have less than $\#s=120$ segments. Then, at each iteration, we 
% use the pairwise score as a filter, 
consider only the pairs that achieve the minimal $T (=4)$ pair dissimilarity and merge the pair which achieves the minimal augmented dissimilarity, 
% starting from a later part of the segmentation process when only 120 segments remain where the remaining segments are more well defined and better separated. 
%The new score combining both scores calculated is:
\begin{align}
\begin{split}
  Aug\_Dissim = Pair\_Dissim - 0.5 \cdot Sil\_Score.
\end{split}
\end{align}
(Note that high {\em Sil\_Score} is preferred for merging.) We refer to this process as re-ranking.
% Like edge confidence and the Pair\_Score, low  Combined\_Score are preferred for merging. On the other hand, high {\em Sil\_Score} indicates better clustering of the union and is favored for merging. We refer to the combined score calculation as reranking. 
% An example of a candidate merged segment and the union of all its neighboring segments can be seen in \ref{rerankExample}.
%This augmented score is consistent with the intuition that a good segments should be uniform in some sense relative to its neighborhood. 
% As shown in the experiments section (sec. \ref{results}), this method achieved an improvement over the simple greedy merging algorithm.

%todo choose if we use re-ranking and if so mention it by name in 4.3 chapter

\subsection{Parametrizing the hierarchy}

The iterative merging process creates a segmentation hierarchy, parameterized by UCM values that should be monotonic with the iteration number (Section \ref{intro:GenSeg}, \cite{ucmMap}). This is the case with edge confidence \cite{gPbucm, cob}, but not necessarily with our  dissimilarity values; see Fig. 1 in Appendix A. 

Using the iteration number itself \cite{MST2Saliency} is problematic: It is not uniformly correlated with  segmentation quality over different images.
Therefore, controlling the segmentation coarseness by uniform, image-independent thresholding, gives uneven segmentations (and low ODS F-score). We used a monotonized value  of the {\em Pair\_Dissim}, and got excellent results. See Appendix A.

\begin{figure*}[th!]
\begin{subfigure}{0.35\textwidth}
\title{BSDS500 - Boundary Measure ($F_{b}$)}
\centering
  \includegraphics[width=0.9\textwidth]{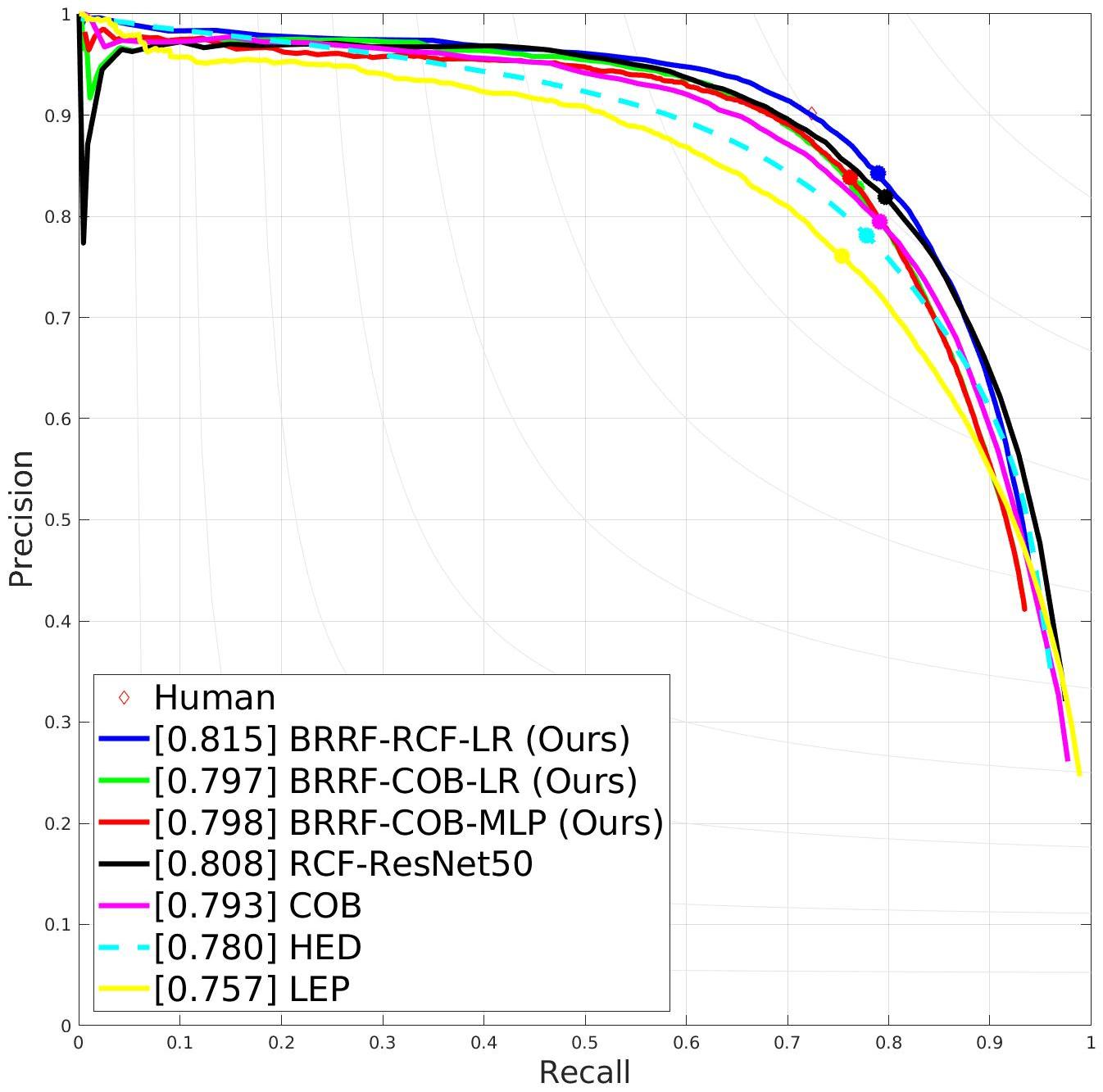}
\end{subfigure}
\begin{subfigure}{0.35\textwidth}
\title{BSDS500 - Region Measure ($F_{op}$)}
\centering
  \includegraphics[width=0.9\textwidth]{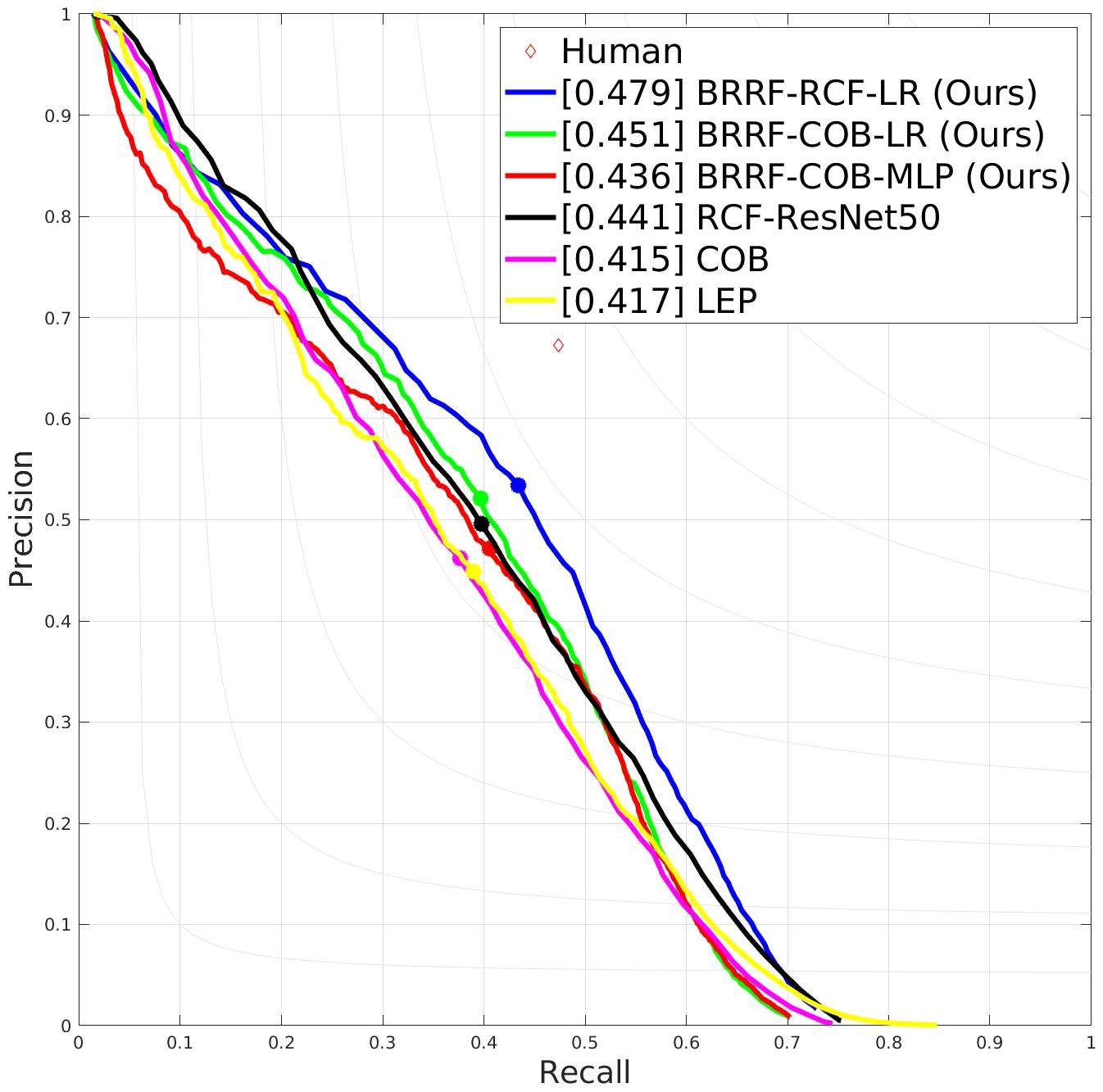}
\end{subfigure}
\begin{subfigure}{0.28\textwidth}
\centering
\scriptsize
\begin{tabular}[t]{lccc}
\hline\noalign{\smallskip}
&& $F_b$ &  \\
Method & ODS & OIS & AP \\
\noalign{\smallskip}
\hline
\noalign{\smallskip}
BRRF-RCF-LR (Ours) & \textbf{0.815} & \textbf{0.838} & 0.855 \\
BRRF-COB-LR (Ours) & 0.797 & 0.820 & 0.835 \\
BRRF-COB-MLP (Ours) & 0.798 & 0.819 & 0.832\\
RCF-ResNet50 \cite{rcf-19} & 0.808 & 0.833 & \textbf{0.873} \\
COB \cite{cob} & 0.793 & 0.820 & 0.859\\
HED \cite{hed} & 0.780 & 0.796 & 0.834 \\
LEP \cite{lep} & 0.757 & 0.793 & 0.828 \\
%MCG \cite{mcg} & 0.747 & 0.779 & 0.759 \\
\noalign{\smallskip}
\hline
\hline\noalign{\smallskip}
&& $F_{op}$ &  \\
Method & ODS & OIS & AP \\
\noalign{\smallskip}
\hline
\noalign{\smallskip}
BRRF-RCF-LR (Ours) & \textbf{0.479} & \textbf{0.531} & \textbf{0.390}\\
BRRF-COB-LR (Ours) & 0.451 & 0.513 & 0.359 \\
BRRF-COB-MLP (Ours) &  0.436 & 0.503 & 0.338\\
RCF-ResNet50 \cite{rcf-19} & 0.441 & 0.5 & 0.370 \\
COB \cite{cob} & 0.415 & 0.466 & 0.333 \\
LEP \cite{lep} & 0.417 & 0.468 & 0.334 \\
%MCG \cite{mcg} & 0.380 & 0.433 & 0.271 \\
\noalign{\smallskip}
\hline
\end{tabular}
\end{subfigure}
\caption{BSDS Test evaluation: Precision-recall curves for evaluation of boundaries \cite{boundaries}, and regions \cite{segmentation_eval}.  Open contour
methods in dashed lines and closed boundaries (from segmentation) in solid lines. ODS, OIS, and AP summary measures. Markers indicate the
optimal operating point, where Fb and Fop are maximized.}\label{BSDS_res}
\end{figure*}

\section{The Pair Dissimilarity Classifier}\label{classifier}
\subsection{Pair Dissimilarity features}\label{class-feat}

We use three types of features for our classification task:
\begin{description}
\item [Representation-based features - ]
The first type is 10 features calculated from the representations of the two associated segments. We use only a subset of pixel representations from the segments (see Section \ref{alg-filt}). For every pair of pixels, one from each segment, we calculate the minimum, maximum, average, and median of both the L2 and the cosine distances between them (following \cite{feat_sign}), as well as distances between the calculated mean of the representation in each segment. The L2 distances are replaced by their log values, to bring all distance calculated to the same range. This choice of features follows classic agglomerative clustering methods \cite{AggClusteringMethod-73}.
% L2 distances We use various distance measurements (TODO: need to cite?) to calculate the following features:

% Following the advice in \cite{feat_sign}, we use both euclidean and cosine distance metrics. 
%The euclidean values undergo a log function to remove scale differences resulting from very large distance values. 
% Due to the potential for large number of pixels in each segment, calculating each pixel pair distance would be expensive and vulnerable to outliers for some of the features. Due to that reason, we filter and sample the representation pixels, the process is elaborated in \ref{alg-comb}.

\item [Edge-based feature - ] The averaged edge probability scores between the segments from an edge detection algorithm \cite{cob,rcf-19}.

\item[Geometric and raw color features - ] The first 3 features describe the geometry of the merged segments: the (square root of the) length of the boundary between the segments, the (square root of the) combined segment size, and the maximum ratio of the boundary (between the segments) length and the segment's  perimeter. 

The last 3 features describe the color dissimilarity and are specified as the 3 differences between the averages of the LAB color space values in the two segments. 
\end{description}

These 17 features serve as input to the classifier that outputs the $Pair\_Dissim$.
While edge detection is very informative, 
% based methods achieve excellent results (e.g.  \cite{cob}), 
our representations encode information about the properties associated with high-level segment content, hence it further boosts performance. 

% Current approaches tend to prefer features that are learned from raw data. The proposed representation follows this preference. The size of the annotated segmentation databases is relatively small, however, and adding some  traditionally specified features helps, possibly by, say, taking a segment's size and shape into account (see  \cite{felzhutt}).

\subsection{Training the classifier}\label{class-train}
To train the classifier, we use segment pairs taken from the segmentation hierarchies generated by the used edge detection algorithm. From each hierarchy, we first generated several segmentations by thresholding the UCM with different values. We use all the neighboring segment pairs in a segmentation as either positive (i.e., should merge) or negative (should not merge) examples. A pair is considered negative when at least 0.6 of its shared boundary is close (within 2 pixels) to the boundary specified in at least one ground truth annotation.
%\cite{BSDS where they refer to the annotation}.
%
The threshold separating between positive and negative examples was empirically determined. 
% Being smaller than 1, it reflects however the preference to false negative examples  over false positive ones, to which the single linkage agglomeration is much more sensitive. 

We experimented with several classifiers, 
% to estimate the {\em Pair \_ Dissim}. 
and found that the best results were obtained with LR (Logistic Regression) and MLP (Multilayer Perceptron); see Appendix C. 

\subsection{Segment filtering and sampling}\label{alg-filt}
The representation is not completely uniform over each segment, and, in particular, includes outliers. 
% , due to the existence of small structures and proximity to different objects. 
Therefore, before using the representation vectors,  we filter them using Isolation Forest \cite{isolationForest}.
For computational efficiency, we also sample large segments ($>300$ pixels). Both filtering and sampling improve the accuracy; see details in Appendix B.
% Once we have the representation pixels, we calculate for every super-pixel pair its likelihood of belonging to the same semantic object. We then proceed to combine the two segments with the highest likelihood, using the filtering and sampling on the new segment and calculating its own likelihood of belonging to the same semantic object with its neighbors. The process of learning an Isolation Forest for each segment is done so, as our segments become larger, we can have more accurate calculation of the representation of the segments due to having more samples, this allows us to be resistant to new segments that contain noisy representation pixels. 

\begin{figure*}[th!]
\begin{subfigure}{0.34\textwidth}
\title{Pascal Context - Boundary Measure ($F_{b}$)}
\centering
  \includegraphics[width=0.89\textwidth]{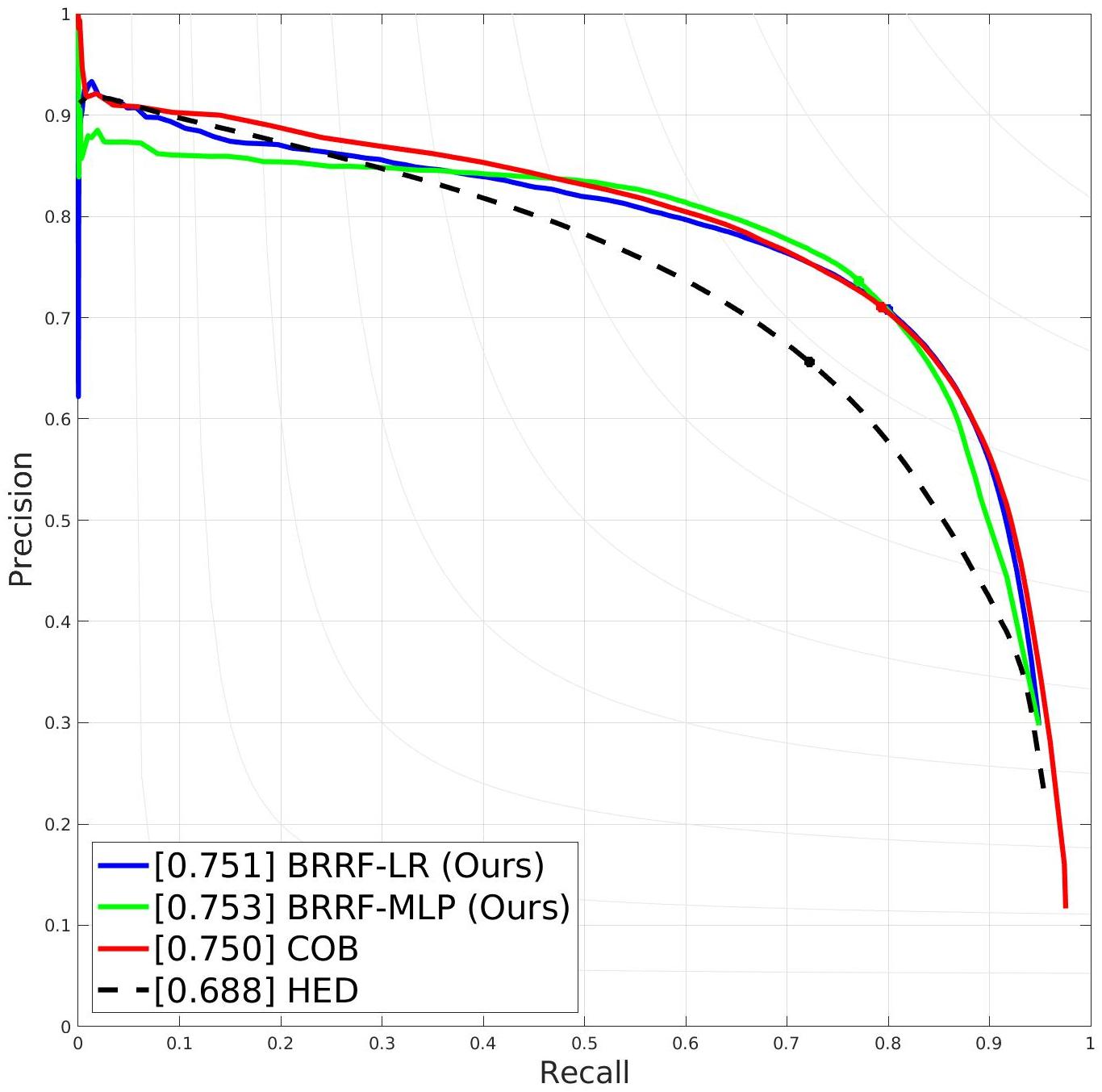}
\end{subfigure}
\begin{subfigure}{0.34\textwidth}
\title{Pascal Context - Region Measure ($F_{op}$)}
\centering
% 1.02
  \includegraphics[width=0.89\textwidth]{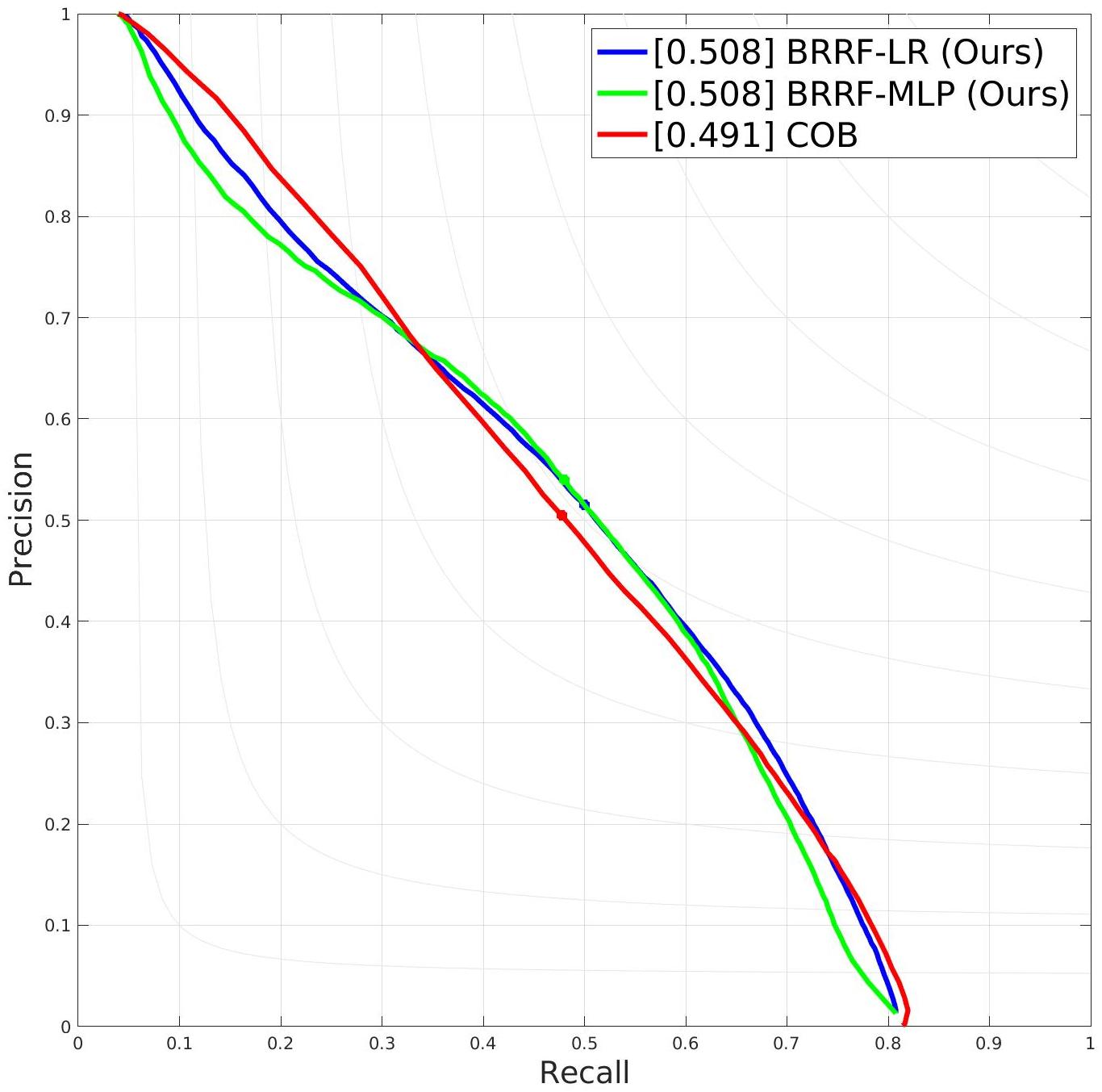}
\end{subfigure}
\begin{subfigure}{0.28\textwidth}
\centering
\scriptsize
\begin{tabular}[t]{lccc}
\hline\noalign{\smallskip}
&& $F_b$ &  \\
Method & ODS & OIS & AP \\
\noalign{\smallskip}
\hline
\noalign{\smallskip}
BRRF-COB-LR (Ours) & 0.751 & \textbf{0.785} & 0.756 \\
BRRF-COB-MLP (Ours) & \textbf{0.753} & 0.781 & 0.750\\
COB \cite{cob} & 0.750 & 0.781 & \textbf{0.773}\\
HED \cite{hed} & 0.688 & 0.707 & 0.704 \\

\noalign{\smallskip}
\hline
\hline\noalign{\smallskip}
&& $F_{op}$ &  \\
Method & ODS & OIS & AP \\
\noalign{\smallskip}
\hline
\noalign{\smallskip}
BRRF-COB-LR (Ours) & \textbf{0.508} & \textbf{0.580} & \textbf{0.439}\\
BRRF-COB-MLP (Ours) &  \textbf{0.508} & 0.579 & 0.427\\
COB \cite{cob} & 0.491 & 0.565 & \textbf{0.439} \\
\noalign{\smallskip}
\hline
\end{tabular}
\end{subfigure}

\caption{Pascal Context Test evaluation: Precision-recall curves for evaluation of boundaries \cite{boundaries}, and regions \cite{segmentation_eval}.} \label{PASCON_res}
\end{figure*}

\section{Experiments}
\label{results}

We first present details of the representation learning procedure and the evaluation of these representations in a pixel pair classification task. We then briefly compare some versions of the agglomerative segmentation algorithm. Finally, we present quantitative and qualitative segmentation results.
%The BRRF algorithm is implemented in Python, and run 45 seconds per image (average). No attempt to optimize the code was done.
The BRRF algorithm runtime is 45 seconds per image (average). No attempt to optimize the code was done.
Our code for both the RepNet and the BRRF algorithm is available on \href{https://github.com/oranshayer/BRRF}{https://github.com/oranshayer/BRRF}.

%Segmentations are obtained when thresholding at optimal dataset scale (ODS) according to the $F_b$ measures for DGS-unary and DGS-SPTs, and according to the $F_{op}$ measures for DGS-DT.

\subsection{RepNet training details}\label{rep-results}

We trained the representation network over the classification task described in Section \ref{Learning-rep} both for the Pascal Context dataset \cite{pascal_context} and BSDS500 dataset \cite{bsds}. For Pascal Context, we started with ImageNet pre-trained weights, and trained with 1000 images containing $5540$ segments. We trained the network for $300$ epochs with $d=8$ and $w_b=5$, SGD optimizer with momentum set to $0.9$, and a learning rate of $0.01$ which was divided by $10$ after $250$ epochs. For the BSDS500 dataset, the network was fine-tuned from the network trained on Pascal, with $300$ \textit{trainval} images containing $2060$ segments. We started with a learning rate of $0.001$ and trained for $300$ epochs.% and achieved $95.2\%$ training accuracy.

\begin{figure*}[h]
\centering
\begin{subfigure}[b]{0.2\textwidth}
  \begin{center}
  \textbf{Image}\par\medskip
  \end{center}
\end{subfigure}
\begin{subfigure}[b]{0.19\textwidth}
  \begin{center}
  \textbf{Representation}\par\medskip
  \end{center}
\end{subfigure}
\begin{subfigure}[b]{0.19\textwidth}
  \begin{center}
  \textbf{BRRF (ours)}\par\medskip
  \end{center}
\end{subfigure}
\begin{subfigure}[b]{0.19\textwidth}
  \begin{center}
  \textbf{RCF \cite{rcf-19}}\par\medskip
  \end{center}
\end{subfigure}
\begin{subfigure}[b]{0.19\textwidth}
  \begin{center}
  \textbf{COB \cite{cob}}\par\medskip
  \end{center}
\end{subfigure}

%\vspace{1 mm}

\begin{subfigure}[b]{0.2\textwidth}
  \includegraphics[width=\textwidth]{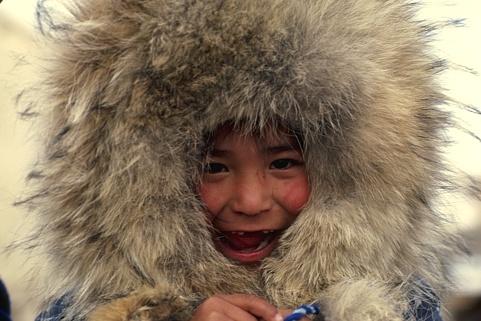}
\end{subfigure}
\begin{subfigure}[b]{0.19\textwidth}
  \includegraphics[width=\textwidth]{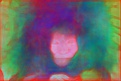}
\end{subfigure}
\begin{subfigure}[b]{0.19\textwidth}
  \includegraphics[width=\textwidth]{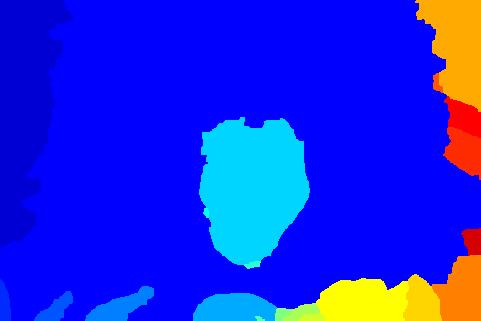}
\end{subfigure}
\begin{subfigure}[b]{0.19\textwidth}
  \includegraphics[width=\textwidth]{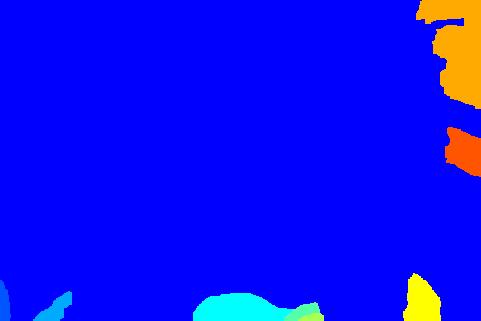}
\end{subfigure}
\begin{subfigure}[b]{0.19\textwidth}
  \includegraphics[width=\textwidth]{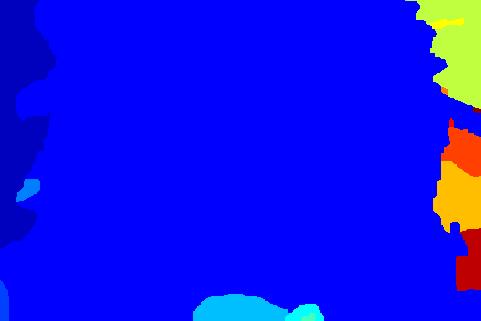}
\end{subfigure}

\vspace{1 mm}
\begin{subfigure}[b]{0.2\textwidth}
  \includegraphics[width=\textwidth]{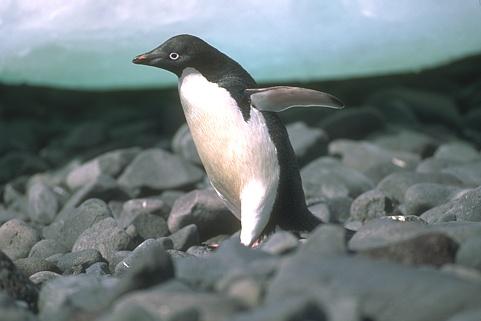}
\end{subfigure}
\begin{subfigure}[b]{0.19\textwidth}
  \includegraphics[width=\textwidth]{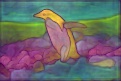}
\end{subfigure}
\begin{subfigure}[b]{0.19\textwidth}
  \includegraphics[width=\textwidth]{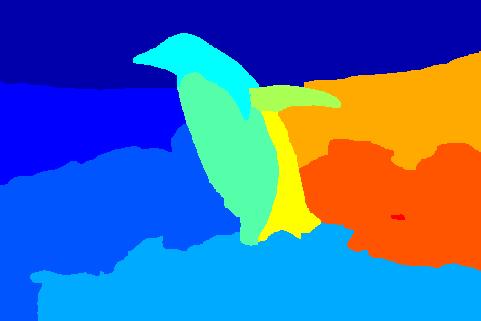}
\end{subfigure}
\begin{subfigure}[b]{0.19\textwidth}
  \includegraphics[width=\textwidth]{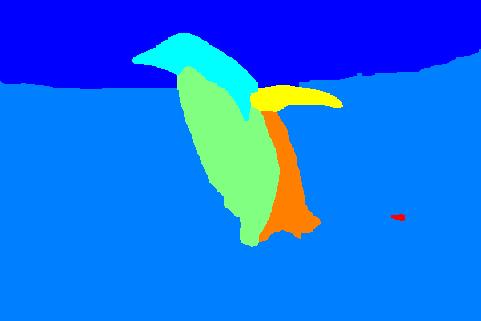}
\end{subfigure}
\begin{subfigure}[b]{0.19\textwidth}
  \includegraphics[width=\textwidth]{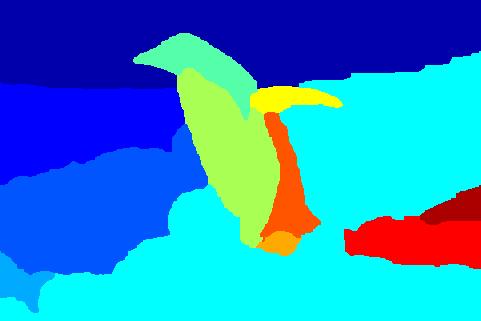}
\end{subfigure}

\vspace{1 mm}
\begin{subfigure}[b]{0.2\textwidth}
  \includegraphics[width=\textwidth]{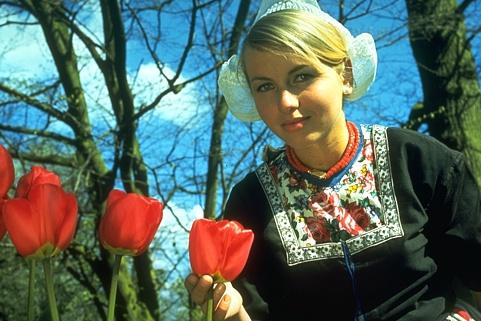}
\end{subfigure}
\begin{subfigure}[b]{0.19\textwidth}
  \includegraphics[width=\textwidth]{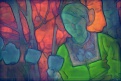}
\end{subfigure}
\begin{subfigure}[b]{0.19\textwidth}
  \includegraphics[width=\textwidth]{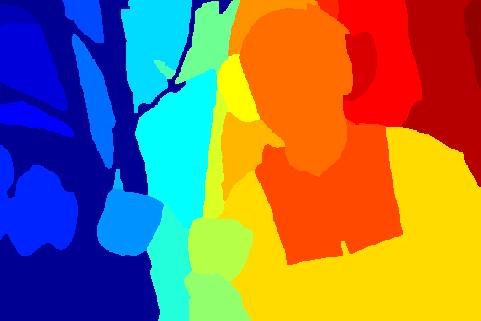}
\end{subfigure}
\begin{subfigure}[b]{0.19\textwidth}
  \includegraphics[width=\textwidth]{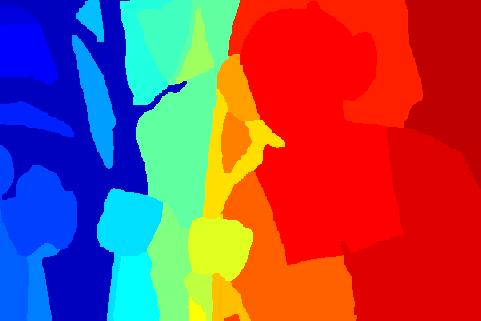}
\end{subfigure}
\begin{subfigure}[b]{0.19\textwidth}
  \includegraphics[width=\textwidth]{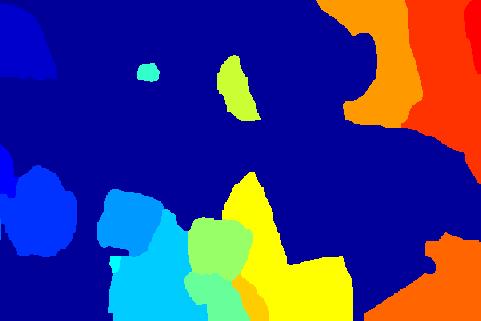}
\end{subfigure}

%\vspace{1 mm}
%\begin{subfigure}[b]{0.2\textwidth}
%  \includegraphics[width=\textwidth]{visual_compare/im_87015.jpg}
%\end{subfigure}
%\begin{subfigure}[b]{0.19\textwidth}
%  \includegraphics[width=\textwidth]{visual_compare/rep_87015.jpg}
%\end{subfigure}
%\begin{subfigure}[b]{0.19\textwidth}
%  \includegraphics[width=\textwidth]{visual_compare/brrf_87015.jpg}
%\end{subfigure}
%\begin{subfigure}[b]{0.19\textwidth}
%  \includegraphics[width=\textwidth]{visual_compare/rcf_87015.jpg}
%\end{subfigure}
%\begin{subfigure}[b]{0.19\textwidth}
%  \includegraphics[width=\textwidth]{visual_compare/cob_87015.jpg}
%\end{subfigure}

\vspace{1 mm}
\begin{subfigure}[b]{0.2\textwidth}
  \includegraphics[width=\textwidth]{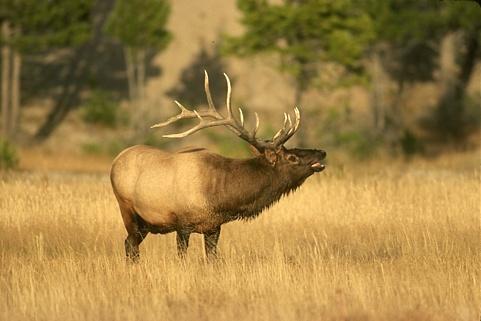}
\end{subfigure}
\begin{subfigure}[b]{0.19\textwidth}
  \includegraphics[width=\textwidth]{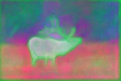}
\end{subfigure}
\begin{subfigure}[b]{0.19\textwidth}
  \includegraphics[width=\textwidth]{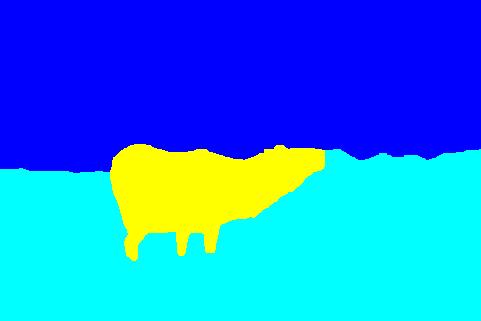}
\end{subfigure}
\begin{subfigure}[b]{0.19\textwidth}
  \includegraphics[width=\textwidth]{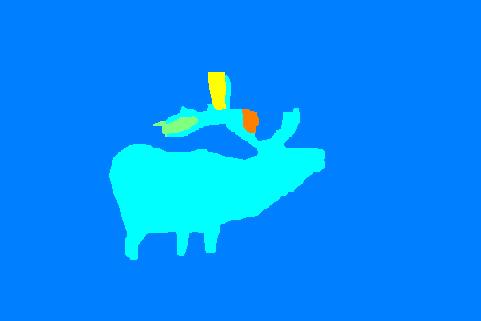}
\end{subfigure}
\begin{subfigure}[b]{0.19\textwidth}
  \includegraphics[width=\textwidth]{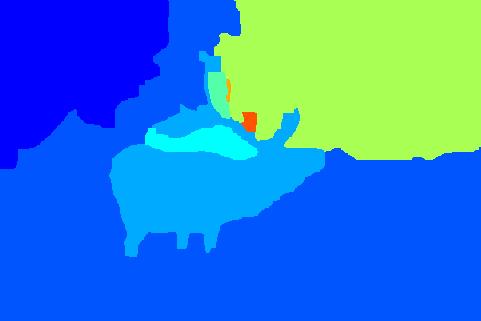}
\end{subfigure}
\caption{Some segmentation examples produced by our algorithms and other competitive algorithms. The segmentations were chosen by maximizing the $F_{op}$ score in the hierarchy. Our algorithm achieves a clearer segmentation between different objects (e.g., the  faces in rows 1 and 3); captures differences between segments of the same class (background in row 2); but struggles with narrow segments (the horns in row 4).}
\end{figure*}

\subsection{Evaluating the representations}\label{rep-results}

To evaluate the quality of the representations themselves, we consider a pixel classification task.
The task is to determine whether two pixels belong to the same segment using the representations. To this end, we use a simple classifier which decides that the pixels belong to the same segment if the Euclidean distance between the representations is smaller than a threshold. The optimal threshold is learned over a validation set. 
The results are presented in Table \ref{table:pixels}.

We compared the representations obtained from our implicit learning method with representations learned as follows: triplet loss, representation from a network trained for semantic segmentation \cite{deeplab}, and representations from a network trained for material classification where materials hold strong texture characteristics \cite{minc}. We also compare with the following pixel representations: RGB, L*a*b, and Gabor filters.
Clearly, our proposed representation achieves the best result, suggesting it can be beneficial for pixel separation tasks such as image segmentation.

\setlength{\tabcolsep}{4pt}
\begin{table}[!h]
\begin{center}
\begin{tabular}{lcc}
\hline\noalign{\smallskip}
Representation & Test accuracy \\
\noalign{\smallskip}
\hline
\noalign{\smallskip}
Gabor filters & 56.09\% \\
RGB & 57.67\% \\
L*a*b & 58.25\% \\
Material classification net \cite{minc} & 70.14\% \\
DeepLab \cite{deeplab} & 71.94\% \\
Triplet loss & 76.34\% \\
Ours (implicit learning) & \textbf{81.04\%} \\
\hline
\end{tabular}
\end{center}
\caption{Pixel pair classification results}\label{table:pixels}
\end{table}
\setlength{\tabcolsep}{1.4pt}

\subsection{Comparing different merging functions}\label{exp:features}

We experimented with several versions of the merging function and found it fairly robust to its parameters. See the more detailed description in Appendix C.

\begin{description}
\item[Different components of the pair dissimilarity] - Table \ref{table:featureEffects} shows the OIS measure of the $F_{op}$ score. For each combination, the classifier was retrained, and then tested on the BSDS500 validation set. Adding the representation noticeably improves our score.
% We did not use re-ranking in the tests.  

% \subsection{Comparing different merging function}\label{exp:features}
% We experimented with several versions of the merging function and found it fairly robust to its parameters. See the full description in App. B. % \begin{description}
\item[Re-ranking] -
% Recall that re-ranking starts when the number of segments is $\#s$ or lower, and is calculated for $T$ candidates (in each iterations). 
Re-ranking required very little additional runtime, but  improved accuracy (see Table \ref{table:featureEffects} and Appendix C).
% (OIS $F_{op}$ increases by $0.003$). 
Increasing $(\#s, T)$ beyond % $\#s=120, T=4$ 
$(120,4)$ 
did not result in further improvement.
\item[Filtering and sampling] - Filtering with the isolation filter improves accuracy (OIS $F_{op}$ increases by 0.004) and increases runtime by 20\%. Using less samples significantly speeds the algorithm and, somewhat surprisingly, slightly increases the accuracy. 
% by reducing the chance of selecting pixels that are not typical to the segment. 
% \item[Choice of Dissimilarity classifiers] We experimented with several classifiers to estimate the {\em Pair \_ Dissim}. The best results were obtained with logistic regression and multilayer perceptron; see  appendix B. 

% \cite{RandomForest}. We optimized the parameters of each classifier and used its probability prediction output as as the Pair Score (for the random forest we used the average of samples on a leaf). Table \ref{table:func_exp} describes the results. 

\end{description}

\subsection{Generic image segmentation}
We evaluated our segmentation algorithm on both the BSDS500 dataset \cite{bsds} and the Pascal Context dataset \cite{pascal_context} (independently) for both LR and MLP classifiers using \cite{cob} for edge detection (denoted as BRRF-COB-LR and BRRF-COB-MLP, respectively). Additionally, we evaluated our algorithm on BSDS500 using \cite{rcf-19} on a ResNet50 architecture for edge detection with an LR classifier (denoted as BRRF-CRF-LR). We trained both the network and merging classifier on the trainval sets on both datasets. \\
\textbf{BSDS500:} Results are presented in Fig. \ref{BSDS_res}.
Using an LR classifier and the \cite{rcf-19} edge detection, we have either matched or improved the state-of-the-art results for the $F_b$ score \cite{boundaries}. Noticeably, we have achieved recall and precision scores that match those of a human annotator. For the $F_{op}$ score \cite{segmentation_eval}, we significantly improved the state of the art for both versions of the F-score and improved the average precision.\\
% The results for the MLP classifier were slightly worse for $F_{op}$ score.\\
%
\textbf{Pascal Context:} 
% Results are presented in fig. \ref{PASCON_res}. 
The results are similar; see Fig. \ref{PASCON_res}. Using an LR classifier, we have either matched or improved the state-of-the-art results on both versions (OIS and ODS) of $F_b$ and $F_{op}$, with more significant improvement of the latter. \\
% The results for the MLP classifier are slightly worse for some of the measures.

For both datasets, the proposed algorithm achieves a more substantial improvement on the region evaluation measures ($F_{op}$). This is expected because our algorithm 
differs by using region representations. While we get the best results in nearly all the $F_{b}$, $F_{op}$ scores, this is not always the case with the AP score, which is influenced also by extreme oversegmentation and undersegmentation 
% uses not only edge properties, but also the learned region representations. 
%as well as the re-ranking stage, both of which rely more on segment context properties, rather than being purely edge reliant.

% The single measure that our algorithm did not outperform the state-of-the-art is the Average Precision (AP) for boundaries, which is calculated by the average precision values along the precision-recall curve. The reason for this under-performance is the initial process of combining small SP to larger segments that causes our algorithm to start from a lower precision compared to other algorithms. Although as can be seen from the precision-recall graph, we have superior values of precision for recall values below 0.85.

\begin{table}[t]
\begin{center}
 \begin{tabularx}{0.4\textwidth}{|c | c | c | c  || X |}
 \hline
  Edge & Geo. and color & Rep. & Re-ranking & $F_{op}$ \\
 \hline \hline
 x & & & & 0.453 \\
  \hline
  x & x & & & 0.484 \\
  \hline
  x &  & x & & 0.49 \\
  \hline
  x &  x & x & & 0.509 \\
  \hline
  x &  x & x & x & 0.512 \\
  \hline
\end{tabularx}
\caption{Features effect on the segmentation results}\label{table:featureEffects}
\end{center}
\end{table}

\section{Conclusion}
\label{chap:conclusion}
We proposed a new approach to generic image segmentation that combines the strengths of edge detection and (a new) pixel-wise representation. The representation, learned through a formulation of a suitable supervised learning task, is a region representation. As such, it complements edge information and improves the segmentation quality, especially when it comes to region-based quality measures. 

Experimentally, the proposed representation, by itself, achieves excellent, state-of-the-art results in pixel pair classification, and overcomes earlier learned and non-learned representations, serving as evidence that it captures characteristics that distinguish between different segments and generalizes well for segments not seen in the training set. The use of these representations through a complete segmentation algorithm
yields state-of-the-art results for generic segmentation.

\vspace{1cm}
{\small
\bibliographystyle{ieee}
\bibliography{egbib}

\begin{thebibliography}{10}\itemsep=-1pt

\bibitem{ucmMap}
P.~Arbelaez.
\newblock Boundary extraction in natural images using ultrametric contour maps.
\newblock In {\em 2006 Conference on Computer Vision and Pattern Recognition
  Workshop (CVPRW'06)}, pages 182--182. IEEE, 2006.

\bibitem{gPbucm}
P.~Arbelaez, M.~Maire, C.~Fowlkes, and J.~Malik.
\newblock Contour detection and hierarchical image segmentation.
\newblock {\em IEEE Trans. Pattern Anal. Mach. Intell.}, 33(5):898--916, May
  2011.

\bibitem{minc}
S.~Bell, P.~Upchurch, N.~Snavely, and K.~Bala.
\newblock Material recognition in the wild with the materials in context
  database.
\newblock {\em Computer Vision and Pattern Recognition (CVPR)}, 2015.

\bibitem{bengio2013representation}
Y.~Bengio, A.~Courville, and P.~Vincent.
\newblock Representation learning: A review and new perspectives.
\newblock {\em IEEE transactions on pattern analysis and machine intelligence},
  35(8):1798--1828, 2013.

\bibitem{minimal_cut-boykov}
Y.~Boykov, O.~Veksler, and R.~Zabih.
\newblock Fast approximate energy minimization via graph cuts.
\newblock {\em IEEE Transactions on pattern analysis and machine intelligence},
  23(11):1222--1239, 2001.

\bibitem{deeplab}
L.-C. Chen, G.~Papandreou, I.~Kokkinos, K.~Murphy, and A.~L. Yuille.
\newblock Semantic image segmentation with deep convolutional nets and fully
  connected crfs.
\newblock In {\em ICLR}, 2015.

\bibitem{siamese2005}
S.~Chopra, R.~Hadsell, and Y.~LeCun.
\newblock Learning a similarity metric discriminatively, with application to
  face verification.
\newblock In {\em Proceedings of the 2005 IEEE Computer Society Conference on
  Computer Vision and Pattern Recognition (CVPR'05) - Volume 1 - Volume 01},
  CVPR '05, pages 539--546, 2005.

\bibitem{meanshift}
D.~Comaniciu, P.~Meer, and S.~Member.
\newblock Mean shift: A robust approach toward feature space analysis.
\newblock volume~24, pages 603--619, 2002.

\bibitem{MST2Saliency}
J.~Cousty, L.~Najman, Y.~Kenmochi, and S.~Guimar{\~a}es.
\newblock Hierarchical segmentations with graphs: quasi-flat zones, minimum
  spanning trees, and saliency maps.
\newblock {\em Journal of Mathematical Imaging and Vision}, 60(4):479--502,
  2018.

\bibitem{decaf}
J.~Donahue, Y.~Jia, O.~Vinyals, J.~Hoffman, N.~Zhang, E.~Tzeng, and T.~Darrell.
\newblock Decaf: A deep convolutional activation feature for generic visual
  recognition.
\newblock In {\em International conference on machine learning}, pages
  647--655, 2014.

\bibitem{felzhutt}
P.~F. Felzenszwalb and D.~P. Huttenlocher.
\newblock Efficient graph-based image segmentation.
\newblock {\em Int. J. Comput. Vision}, 59(2):167--181, Sept. 2004.

\bibitem{p2v}
O.~Fried, S.~Avidan, and D.~Cohen-Or.
\newblock Patch2vec: Globally consistent image patch representation.
\newblock In {\em Computer Graphics Forum}, volume~36, pages 183--194. Wiley
  Online Library, 2017.

\bibitem{siamese2006}
R.~Hadsell, S.~Chopra, and Y.~LeCun.
\newblock Dimensionality reduction by learning an invariant mapping.
\newblock In {\em Computer vision and pattern recognition, 2006 IEEE computer
  society conference on}, volume~2, pages 1735--1742. IEEE, 2006.

\bibitem{resnet}
K.~He, X.~Zhang, S.~Ren, and J.~Sun.
\newblock Deep residual learning for image recognition.
\newblock {\em arXiv preprint arXiv:1512.03385}, 2015.

\bibitem{hoffer2015triplet}
E.~Hoffer and N.~Ailon.
\newblock Deep metric learning using triplet network.
\newblock In {\em International Workshop on Similarity-Based Pattern
  Recognition}, pages 84--92. Springer, 2015.

\bibitem{feat_sign}
S.~Horiguchi, D.~Ikami, and K.~Aizawa.
\newblock Significance of softmax-based features in comparison to distance
  metric learning-based features.
\newblock {\em IEEE transactions on pattern analysis and machine intelligence},
  2019.

\bibitem{koch15}
G.~Koch, R.~Zemel, and R.~Salakhutdinov.
\newblock Siamese neural networks for one-shot image recognition.
\newblock In {\em ICML Deep Learning Workshop}, volume~2, 2015.

\bibitem{isolationForest}
F.~T. Liu, K.~M. Ting, and Z.-H. Zhou.
\newblock Isolation forest.
\newblock In {\em 2008 Eighth IEEE International Conference on Data Mining},
  pages 413--422. IEEE, 2008.

\bibitem{rcf-19}
Y.~Liu, M.-M. Cheng, X.~Hu, J.-W. Bian, L.~Zhang, X.~Bai, and J.~Tang.
\newblock Richer convolutional features for edge detection.
\newblock {\em IEEE Transactions on Pattern Analysis and Machine Intelligence
  (TPAMI)}, 41(8):1939--1946, 2019.

\bibitem{del}
Y.~Liu, P.-T. Jiang, V.~Petrosyan, S.-J. Li, J.~Bian, L.~Zhang, and M.-M.
  Cheng.
\newblock Del: Deep embedding learning for efficient image segmentation.
\newblock In {\em IJCAI}, 2018.

\bibitem{cob}
K.~Maninis, J.~Pont-Tuset, P.~Arbel\'{a}ez, and L.~V. Gool.
\newblock Convolutional oriented boundaries: From image segmentation to
  high-level tasks.
\newblock {\em IEEE Transactions on Pattern Analysis and Machine Intelligence
  (TPAMI)}, 2017.

\bibitem{bsds}
D.~Martin, C.~Fowlkes, D.~Tal, and J.~Malik.
\newblock A database of human segmented natural images and its application to
  evaluating segmentation algorithms and measuring ecological statistics.
\newblock In {\em Proc. 8th Int'l Conf. Computer Vision}, volume~2, pages
  416--423, July 2001.

\bibitem{boundaries}
D.~R. Martin, C.~C. Fowlkes, and J.~Malik.
\newblock Learning to detect natural image boundaries using local brightness,
  color, and texture cues.
\newblock {\em IEEE Trans. Pattern Anal. Mach. Intell.}, 26(5):530--549, May
  2004.

\bibitem{pascal_context}
R.~Mottaghi, X.~Chen, X.~Liu, N.-G. Cho, S.-W. Lee, S.~Fidler, R.~Urtasun, and
  A.~Yuille.
\newblock The role of context for object detection and semantic segmentation in
  the wild.
\newblock In {\em The IEEE Conference on Computer Vision and Pattern
  Recognition (CVPR)}, June 2014.

\bibitem{watershed}
L.~Najman and M.~Schmitt.
\newblock Geodesic saliency of watershed contours and hierarchical
  segmentation.
\newblock {\em IEEE Transactions on pattern analysis and machine intelligence},
  18(12):1163--1173, 1996.

\bibitem{face_veldadi}
O.~M. Parkhi, A.~Vedaldi, A.~Zisserman, et~al.
\newblock Deep face recognition.
\newblock In {\em BMVC}, volume~1, page~6, 2015.

\bibitem{mcg}
J.~Pont-Tuset, P.~Arbel\'{a}ez, J.~Barron, F.~Marques, and J.~Malik.
\newblock Multiscale combinatorial grouping for image segmentation and object
  proposal generation.
\newblock In {\em arXiv:1503.00848}, March 2015.

\bibitem{segmentation_eval}
J.~Pont-Tuset and F.~Marques.
\newblock Supervised evaluation of image segmentation and object proposal
  techniques.
\newblock {\em IEEE Transactions on Pattern Analysis and Machine Intelligence
  (TPAMI)}, 38(7):1465--1478, 2016.

\bibitem{yolo}
J.~Redmon and A.~Farhadi.
\newblock Yolo9000: Better, faster, stronger.
\newblock {\em arXiv preprint arXiv:1612.08242}, 2016.

\bibitem{silhouette}
P.~J. Rousseeuw.
\newblock Silhouettes: a graphical aid to the interpretation and validation of
  cluster analysis.
\newblock {\em Journal of computational and applied mathematics}, 20:53--65,
  1987.

\bibitem{facenet}
F.~Schroff, D.~Kalenichenko, and J.~Philbin.
\newblock Facenet: A unified embedding for face recognition and clustering.
\newblock In {\em CVPR}, pages 815--823. IEEE Computer Society, 2015.

\bibitem{ncuts}
J.~Shi and J.~Malik.
\newblock Normalized cuts and image segmentation.
\newblock {\em IEEE Trans. Pattern Anal. Mach. Intell.}, 22(8):888--905, Aug.
  2000.

\bibitem{AggClusteringMethod-73}
P.~H. Sneath, R.~R. Sokal, et~al.
\newblock {\em Numerical taxonomy. The principles and practice of numerical
  classification.}
\newblock 1973.

\bibitem{deepface}
Y.~Taigman, M.~Yang, M.~Ranzato, and L.~Wolf.
\newblock Deepface: Closing the gap to human-level performance in face
  verification.
\newblock In {\em Proceedings of the 2014 IEEE Conference on Computer Vision
  and Pattern Recognition}, CVPR '14, pages 1701--1708, Washington, DC, USA,
  2014. IEEE Computer Society.

\bibitem{t-SNE}
L.~van~der Maaten and G.~Hinton.
\newblock Visualizing high-dimensional data using t-sne.
\newblock {\em Journal of Machine Learning Research}, 9: 2579–2605, Nov 2008.

\bibitem{hed}
S.~Xie and Z.~Tu.
\newblock Holistically-nested edge detection.
\newblock In {\em Proceedings of IEEE International Conference on Computer
  Vision}, 2015.

\bibitem{zeiler2014visualizing}
M.~D. Zeiler and R.~Fergus.
\newblock Visualizing and understanding convolutional networks.
\newblock In {\em European conference on computer vision}, pages 818--833.
  Springer, 2014.

\bibitem{lep}
Q.~Zhao.
\newblock Segmenting natural images with the least effort as humans.
\newblock In M.~W.~J. Xianghua~Xie and G.~K.~L. Tam, editors, {\em Proceedings
  of the British Machine Vision Conference (BMVC)}, pages 110.1--110.12. BMVA
  Press, September 2015.

\end{thebibliography}
}

\end{document}